%% file: main.tex
\documentclass[10pt,twocolumn,letterpaper]{article}

\usepackage[pagenumbers]{cvpr} %


\definecolor{cvprblue}{rgb}{0.21,0.49,0.74}
\usepackage[pagebackref,breaklinks,colorlinks,allcolors=cvprblue]{hyperref}
\usepackage{natbib}
\newcommand{\dv}{\Delta v}
\usepackage{algorithm}
\def\confName{CVPR}
\def\confYear{2027}
\usepackage{amsmath,amssymb}
\usepackage{graphicx}
\usepackage{subcaption}            %
\usepackage{algorithm}             %
\usepackage{algpseudocode}         %

\usepackage{booktabs}
\usepackage{array} 
\usepackage{pifont}

\newcommand{\vimg}{v^{\mathrm{img}}}
\newcommand{\vvid}{v^{\mathrm{vid}}}
\newcolumntype{L}[1]{>{\raggedright\arraybackslash}p{#1}}

\newcommand{\method}{TOGO}

\title{\method: Take One Image Editor, Get One Video Editor for Free}
\author{Yixuan Luo\footnotemark[1]\\
University of Chicago\\
\and
Jose Caballero\\
Epic Games\\
\and
Yanjing Li\\
Northeastern University\\
}

\begin{document}
\twocolumn[{%
\renewcommand\twocolumn[1][]{#1}%
\maketitle
\begin{center}
\centering
\includegraphics[width=\linewidth]{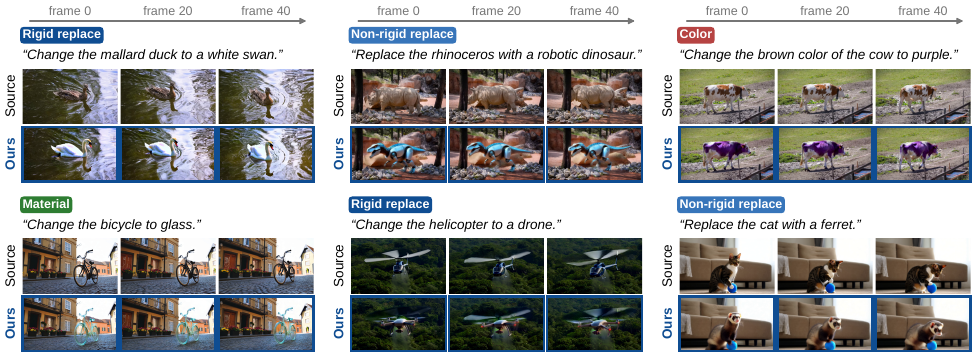}
\captionof{figure}{\textbf{\method{} turns an image editor into a video editor guide without any training.} We add the edit velocity
of an off-the-shelf image editor (Qwen-Image-Edit) to the sampler of a video model (Wan2.1) and confine it to the edited
object. The result follows the instruction consistently over time, while the background, camera, and motion stay faithful
to the source. All examples are from FiVE-Bench; frames 0, 20 and 40 of 41.}
\label{fig:teaser}
\end{center}%
}]
{\renewcommand\thefootnote{\fnsymbol{footnote}}%
\footnotetext[1]{Work done during an internship at Epic Games.}}
\input{sec/abstract}

\input{sec/introduction}
\input{sec/relatedwork}

\input{sec/method}

\input{sec/experiment}

\input{sec/conclusions}

{
    \small
    \bibliographystyle{ieeenat_fullname}
    \bibliography{main}
}

\end{document}

%% file: sec/abstract.tex
\begin{abstract}
Instruction-based image editors are trained on millions of edit pairs and follow instructions far more reliably than
current video editors. We show that this ability can be transferred to video editing
without any training. Our key observation is that a growing number of image editors and video
generators share the same latent space. Their velocity fields can therefore be combined at sampling time: the
difference between the editor's velocity under the instruction and under a neutral prompt, which we call the
\emph{edit velocity}, can be added to the sampler of a video model. Added naively, however, it blurs the video, alters
the background and flickers. An analysis of the edit velocity explains why. It is small and reliable only at coarse
scales and high noise levels, not concentrated on the edited object, and uncorrelated across frames. From these
properties, we derive a simple guidance operator that amplifies the edit velocity and restricts it in noise level,
space, and time. The
video model then turns the incoherent per-frame signal into a temporally coherent edit. Our method, \method{},
combines Qwen-Image-Edit with the Wan2.1 video model without training or fine-tuning. On FiVE-Bench, it raises the
edit success rate from 47.0\% (the same video model without the editor) to 65.6\%, above the best published
method (59.2\%). At the same time, it maintains the background PSNR and obtains the best LPIPS and SSIM of all
methods. Because the video editor is assembled at inference time, it can be upgraded to any newer image editor
that shares the latent space with no cost.
\end{abstract}

%% file: sec/introduction.tex
\section{Introduction}
\label{sec:intro}

Instruction-based image editing has advanced rapidly. Starting from InstructPix2Pix~\citep{brooks2023instructpix2pix}
and culminating in large editors such as Qwen-Image-Edit~\citep{wu2025qwenimage}, these models are trained on millions
of (image, instruction, edited image) triplets. A user can change the color, material or identity of an object with
one sentence, and the rest of the image stays untouched. However, video editing has not kept up. Collecting edited videos is
far more expensive than collecting edited images, and the few video editors trained at scale required building large
datasets first~\citep{bai2026qwenvideoedit}. Training-free video editors instead reuse text-to-image or text-to-video
models~\citep{geyer2024tokenflow,li2024vidtome,yatim2024dmt,yang2025videograin,li2025five}. They infer the edit from
the difference between a source and a target caption, which often leads to weak or wrong changes.

A strong image editor can be applied to every frame, but the result flickers because the editor has no notion of
time~\citep{bai2026qwenvideoedit}. Editing only the first frame and propagating it with an image-to-video
model~\citep{ku2024anyv2v} avoids flicker but often loses the edit: on FiVE-Bench it succeeds on only 38\% of the edits
and performs most poorly in preserving the source motion of all methods (\cref{tab:five}). In neither case do the image editor, which
knows what the edit should look like, and the video model, which knows how a video evolves, work \emph{together}.

We observe that many current generative models share a latent space. Qwen-Image-Edit uses the VAE encoder of the
Wan2.1 video model~\citep{wan2025}, as do Wan2.1-14B, Wan2.2-T2V-A14B, VACE, SkyReels-V2 and LongCat-Video, and all of
them are trained with flow matching~\citep{liu2023flow,lipman2023flow}. Their velocities can therefore be combined at
sampling time, much like the conditional predictions in classifier-free guidance~\citep{ho2022cfg}. We take the
difference between the image editor's velocity~\cite{kulikov2025flowedit} under the instruction and under a neutral prompt, which we call the
\emph{edit velocity}, and add it to the sampler of the video model (\cref{fig:teaser}).

Added naively, the edit velocity blurs the frames, changes the background and makes the edit flicker. Our analysis
(\cref{sec:method:anatomy}) explains why: the edit velocity is small, reliable
only at high noise levels and coarse scales, spread over the whole frame, and uncorrelated across frames. We therefore
amplify it, restrict it to high noise levels, confine it to the edited object and smooth it over neighboring frames.
The video model integrates the remaining per-frame guidance into a temporally coherent edit that keeps the motion and
the background of the source. The resulting method, \method{}, adds no trainable parameters and uses one set of
hyperparameters for all videos and edits.

On FiVE-Bench, \method{} combines Qwen-Image-Edit with Wan2.1 and reaches a FiVE-Acc, the benchmark's VLM-judged edit
success rate, of 65.6\%: 6.4 points above the strongest baseline and 18.6 points above the same video model without
the editor, with the best LPIPS and SSIM of all methods. It also ranks first under a second VLM judge from a different
model family, is preferred in a blind user study (66\% of the comparisons for instruction following, over 70\% for
preservation and temporal stability), and transfers without retuning to a newer image editor and to a $10\times$
larger video model.

In summary, our contributions are:
\begin{itemize}
  \item We first show that an instruction-based image editor can be turned into a video editor without training, by adding
  its edit velocity to the sampler of a video model that shares its latent space.
  \item We analyze the edit velocity, identify why naive guidance fails, and derive a guidance operator that the video
  model turns into a temporally coherent edit.
  \item \method{} sets a new state of the art in edit success on FiVE-Bench while keeping background
  preservation and motion fidelity close to the video model it builds on, without training.
  \item \method{} can be upgraded to any newer image
  editor or video model in the same latent space.
\end{itemize}

%% file: sec/relatedwork.tex
\section{Related Work}
\label{sec:related}

\paragraph{Editing with image diffusion priors.}
Early video editing methods build on text-to-image models and make the sampler aware of neighbouring frames, by fitting temporal attention to the clip~\cite{wu2023tuneavideo} or, without tuning, by sharing attention or propagating features~\cite{ceylan2023pix2video,qi2023fatezero,yang2023rerender,geyer2024tokenflow}; the concurrent EditVid~\cite{juvekar2026editvid} applies this to an MM-DiT editor. Relying on hand-designed propagation, they flicker and drift under large motion.

\paragraph{Editing with video diffusion priors.}
Open video generators~\cite{wan2025} provide a learned temporal prior. Image-editing samplers can be applied to them via SDEdit-style partial noising~\cite{meng2022sdedit} or inversion-free FlowEdit~\cite{kulikov2025flowedit}, and extended with masks, object-confined residuals, or feature injection~\cite{li2026flowdirector,wang2026oavc,bai2025uniedit}. The edit, however, is still carried by text conditioning trained for description rather than instruction following. AnyV2V~\cite{ku2024anyv2v}, Videoshop~\cite{fan2024videoshop} and I2VEdit~\cite{ouyang2024i2vedit} instead edit the first frame with an image editor and propagate it with an image-to-video model. We also take the edit from an image editor, but inject its signal into every high-noise step of a T2V model, so the edit is not tied to one frame and needs no per-video training.

\paragraph{Trained instruction-based editors.}
Trained editors~\cite{zi2025senorita,bai2025ditto,jiang2025vace,bai2026qwenvideoedit} learn instruction following from synthetic video pairs; Qwen-Video-Edit~\cite{bai2026qwenvideoedit} even uses the same two models as we do, fine-tuned on Ditto-1M. They follow instructions well but inherit the habits of their synthetic data, whereas we train nothing and stay on the source video's FlowEdit trajectory, keeping unedited regions close to the input.

\paragraph{Guidance and composing pretrained models.}
Classifier guidance~\cite{dhariwal2021classifier} and classifier-free guidance~\cite{ho2022cfg} add a difference of predictions to each update, and Delta Denoising Score~\cite{Hertz_2023_ICCV} edits images with the difference of two conditional predictions; we apply this to the image editor's velocity. Other works compose pretrained models at inference, sequentially (BIVDiff~\cite{shi2024bivdiff}) or within each sampling step (VideoElevator~\cite{zhang2025videoelevator}, FLDM~\cite{lu2024fldm}, VideoGuide~\cite{lee2025videoguide}). We belong to the second group, but our image editor never produces frames: it only adds a classifier-free-style difference to the video model's velocity, possible because the two models share a latent space. As in VideoGuide, and consistent with VeloEdit's~\cite{li2026veloedit} finding that the first steps fix an edit, we confine this term to the high-noise phase.

%% file: sec/method.tex
\begin{figure*}[t]
\centering
\includegraphics[width=\linewidth]{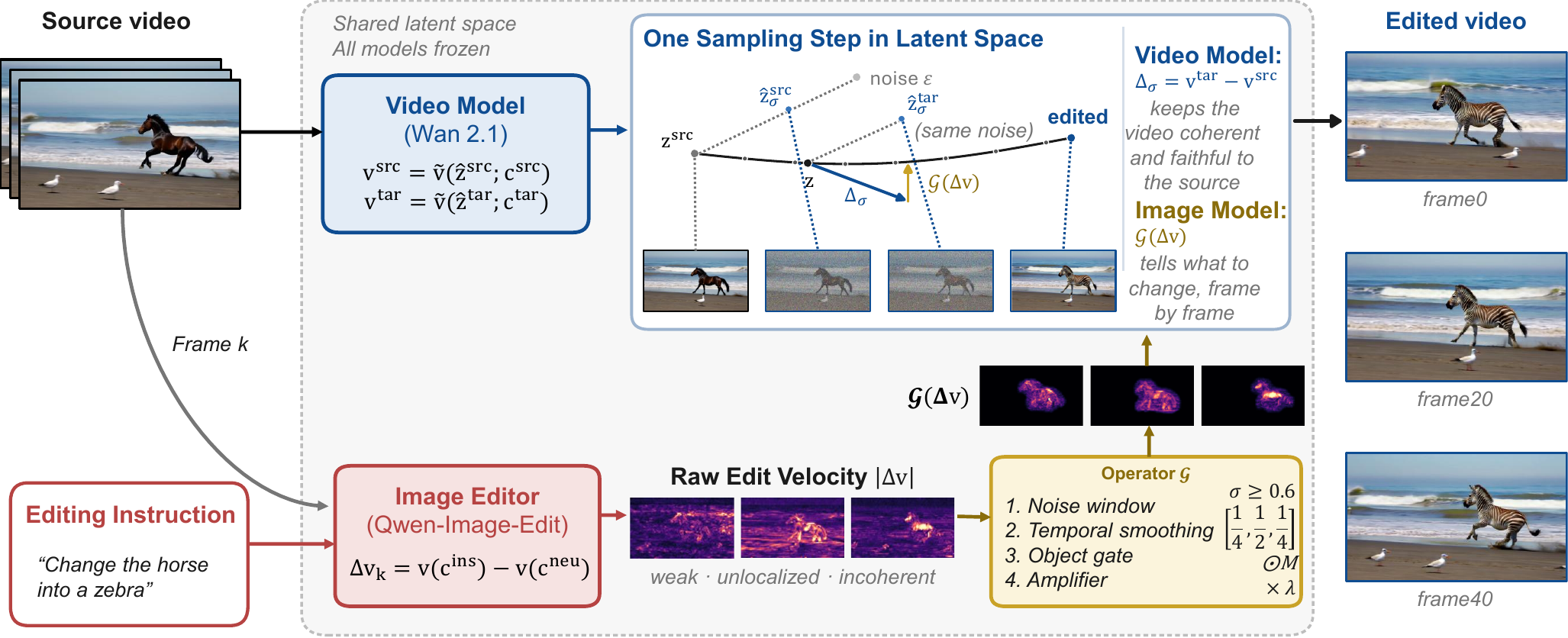}
\caption{\textbf{Overview of \method{}.} Two frozen models that share a latent space are combined at sampling time.
At every step, the video model provides the FlowEdit velocity $\Delta_\sigma$, and the image editor provides, for each
latent frame $k$, the edit velocity $\Delta v_k$ between the instruction and a neutral prompt. The raw $\Delta v$ is
weak, unlocalized and incoherent across frames; $\mathcal{G}$ amplifies, windows, smooths and gates it, and the video
model turns $\mathcal{G}(\Delta v)$ into a coherent edit that keeps the motion and background of the source.}
\label{fig:pipeline}
\end{figure*}
\section{Method}
\label{sec:method}

Given a source video $\mathbf{X}^{\mathrm{src}}$ and an instruction $c^{\mathrm{ins}}$ requesting a local change to one
object, we seek an edited video that applies the change to that object and leaves everything else, including motion,
unchanged. \method{} combines two frozen models that share a latent space (\cref{sec:method:latent}): a text-to-video
generator $\vvid$ that produces the video, and an instruction-based image editor $\vimg$ that specifies the change.
$\vvid$ runs inversion-free FlowEdit sampling~\cite{kulikov2025flowedit} on the source latents, and we add the editor's
\emph{edit velocity} $\dv$ to its update as guidance (\cref{sec:method:dv}). Analyzing $\dv$ reveals a weak but genuine
edit direction buried in spatially diffuse, temporally incoherent components. These findings motivate the four parts of
an operator $\mathcal{G}$ that turns $\dv$ into video guidance: an amplifier, a noise-level window, an object gate, and
temporal smoothing (\cref{sec:method:anatomy}). \Cref{fig:pipeline} gives an overview.

\subsection{Preliminaries}
\label{sec:method:prelim}

\paragraph{Flow matching.}
Both models are trained with flow matching~\cite{lipman2023flow,liu2023flow} on the linear path
\begin{equation}
  \mathbf{x}_\sigma = (1-\sigma)\,\mathbf{x}_0 + \sigma\,\boldsymbol{\epsilon}, \quad \sigma\in[0,1],
  \label{eq:path}
\end{equation}
and regress its velocity $\boldsymbol{\epsilon}-\mathbf{x}_0$ under a condition $c$; $\tilde v_\theta$ denotes the
velocity after classifier-free guidance~\cite{ho2022cfg}. Velocities of networks that share this path live in one
space and can be added or subtracted.

\paragraph{FlowEdit.}
FlowEdit~\cite{kulikov2025flowedit} transports an estimate $\mathbf{z}$, initialized at the source latent
$\mathbf{z}^{\mathrm{src}}$, from a source caption $c^{\mathrm{src}}$ to a target caption $c^{\mathrm{tar}}$ without
inversion. At each step it places the source and the estimate on the path with the \emph{same} noise,
\begin{equation}
  \hat{\mathbf{z}}^{\mathrm{src}}_\sigma = (1-\sigma)\,\mathbf{z}^{\mathrm{src}} + \sigma\boldsymbol{\epsilon},\quad
  \hat{\mathbf{z}}^{\mathrm{tar}}_\sigma = \hat{\mathbf{z}}^{\mathrm{src}}_\sigma + (\mathbf{z} - \mathbf{z}^{\mathrm{src}}),
  \label{eq:coupled_states}
\end{equation}
and integrates the difference of the two conditional velocities,
\begin{equation}
  \begin{split}
  \Delta_\sigma &= \tilde v_\theta(\hat{\mathbf{z}}^{\mathrm{tar}}_\sigma,\sigma,c^{\mathrm{tar}}) - \tilde v_\theta(\hat{\mathbf{z}}^{\mathrm{src}}_\sigma,\sigma,c^{\mathrm{src}}), \\
  \mathbf{z} &\leftarrow \mathbf{z} + (\sigma_{i+1}-\sigma_i)\,\Delta_{\sigma_i}.
  \end{split}
  \label{eq:flowedit}
\end{equation}
Because $\boldsymbol\epsilon$ is shared,
the part of the velocity that reconstructs the source cancels, so the result stays anchored to the input and, inside a
video generator, preserves layout and motion well~\cite{li2025five}. A text-to-video model, however, renders captions
rather than following instructions, so the caption difference yields a weak and imprecise edit. Nothing requires the
edit direction to come from the video model: any network on the same path can supply it. We take it from an
instruction-based image editor.

\subsection{A Shared Latent Space}
\label{sec:method:latent}

Qwen-Image reuses the frozen Wan2.1 VAE~\cite{wu2025qwenimage}: the encoder shipped with Qwen-Image-Edit is a bfloat16
copy of the Wan2.1 encoder and yields identical latents at the same precision. Both models also follow the
parameterization of \cref{eq:path}. A noisy video latent can thus be passed to the image editor at the same noise level
$\sigma$, and both networks return velocities in the same space.

\paragraph{Frame correspondence.}
The video VAE compresses time by $4\times$: $N=4K'+1$ frames become $K=K'+1$ latent frames
$\mathbf{z}=[\mathbf{z}^{(0)},\dots,\mathbf{z}^{(K-1)}]$, where $\mathbf{z}^{(0)}$ encodes RGB frame $0$ and
$\mathbf{z}^{(k)}$, $k\ge1$, summarizes RGB frames $4k-3,\dots,4k$. Since the editor is also conditioned on the image
being edited, we assign each latent frame a representative source frame
\begin{equation}
  \mathbf{I}^{(k)} = \mathbf{X}^{\mathrm{src}}_{f(k)}, \quad f(0)=0,\quad f(k)=4k-1 \;\;(k\ge 1).
  \label{eq:frame_map}
\end{equation}
Latent frames with $k\ge1$ mix four RGB frames and lie outside the editor's training distribution. The operators of \cref{sec:method:consistency} make the guidance robust to it.

\subsection{The Edit Velocity as Guidance}
\label{sec:method:dv}

The image editor's prediction depends on the noisy latent $\mathbf{x}$, the condition image $\mathbf{I}$ and a text
prompt. We define its \emph{edit velocity} as the change in prediction caused by the instruction alone, measured
against a neutral prompt $c^{\mathrm{neu}}$ (``Keep the image exactly as it is.''):
\begin{equation}
  \dv(\mathbf{x},\sigma;\mathbf{I}) = \vimg(\mathbf{x},\sigma \mid \mathbf{I}, c^{\mathrm{ins}}) - \vimg(\mathbf{x},\sigma \mid \mathbf{I}, c^{\mathrm{neu}})
  \label{eq:dv}
\end{equation}
The two terms share everything except the prompt, so $\dv$ isolates the direction in which the editor would move the
sample to realize the edit. It plays the role of the caption difference in \cref{eq:flowedit}, but it comes from a
model trained to follow instructions and is conditioned on the actual frame rather than on a caption.

We evaluate \cref{eq:dv} independently on every latent frame of FlowEdit's target state, with the correspondence of
\cref{eq:frame_map},
\begin{equation}
  \begin{gathered}
  \mathbf{g}^{(k)}_\sigma = \dv\big(\hat{\mathbf{z}}^{\mathrm{tar},(k)}_\sigma,\sigma;\mathbf{I}^{(k)}\big),\quad
  k=0,\dots,K-1.
  \end{gathered}
  \label{eq:g}
\end{equation}
Evaluating on the target state means that the editor sees the current, partially edited estimate, so it keeps pushing
only as long as the edit is incomplete. We stack the results into $\mathbf{g}_\sigma$ and add them to the FlowEdit
update:
\begin{equation}
  \begin{split}
  \mathbf{z} \leftarrow{}& \mathbf{z} + (\sigma_{i+1}-\sigma_i)\big(\Delta_{\sigma_i} + \mathcal{G}(\mathbf{g}_{\sigma_i})\big).
  \end{split}
  \label{eq:update}
\end{equation}
Here $\mathcal{G}$ is an operator that turns the raw per-frame edit velocity into guidance for the video. The roles are
explicit in \cref{eq:update}: $\Delta_\sigma$ keeps the sample coherent and anchored to the source, and
$\mathbf{g}_\sigma$ says what should change. 

\subsection{From Edit Velocity to Video Guidance}
\label{sec:method:anatomy}
\label{sec:method:consistency}

\begin{figure*}[t]
\centering
\includegraphics[width=\linewidth]{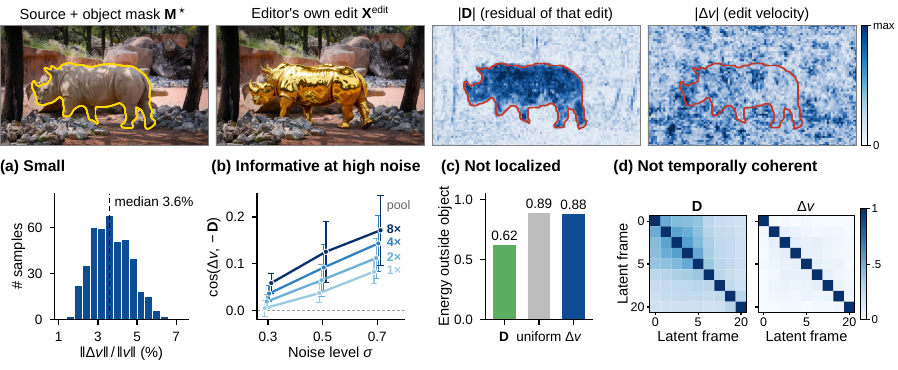}%
{\phantomsubcaption\label{fig:anatomy:a}%
 \phantomsubcaption\label{fig:anatomy:b}%
 \phantomsubcaption\label{fig:anatomy:c}%
 \phantomsubcaption\label{fig:anatomy:d}}%
\caption{\textbf{What the edit velocity $\dv$ contains} (8 local edits from DAVIS-2017~\cite{pont2017davis}, two noise
seeds). \emph{Top:} for ``make the rhino shiny gold'' (latent frame~0), the residual $\mathbf{D}$ of the editor's full
edit concentrates on the object $\mathbf{M}^\star$, whereas $\dv$ spreads over the whole frame.
\textbf{(a)}~$\dv$ is a small fraction of the full velocity.
\textbf{(b)}~Its alignment with the ideal direction $-\mathbf{D}$ (latent frame~0) grows with noise level and spatial
pooling (mean $\pm$ s.e.\ over videos).
\textbf{(c)}~Its energy fraction outside the object (mean over videos) matches that of a spatially uniform signal.
\textbf{(d)}~Cross-frame cosine similarity at $\sigma=0.7$ (latent frames 0--20, sampled): $\mathbf{D}$ is shared across
frames; $\dv$ is not.}
\label{fig:anatomy}
\end{figure*}

A shared latent space makes $\dv$ \emph{addable} to the video velocity; it does not make it \emph{useful} as is. We
analyze it on 8 videos with one local edit each, used only to design the method (\cref{sec:exp:setup}), at several
noise levels and with two noise seeds. As the ideal edit direction, we take the latent residual of the editor's
\emph{own} complete edit, $\mathbf{D}^{(k)}=\mathcal{E}(\mathbf{X}^{\mathrm{edit}})^{(k)}-\mathbf{z}^{\mathrm{src},(k)}$,
where $\mathbf{X}^{\mathrm{edit}}$ edits every frame independently with the full editor. Since
$v=\boldsymbol\epsilon-\mathbf{x}_0$, moving the clean sample by $\mathbf{D}$ at fixed noise changes the velocity by
$-\mathbf{D}$, so a useful $\dv$ should point along $-\mathbf{D}$. \Cref{fig:anatomy} summarizes four findings. Each
motivates one deliberately simple part of $\mathcal{G}$, and everything else is left to the video model.

\paragraph{(a) The signal is small $\rightarrow$ amplifier.}
The predictions with and without the instruction have a cosine similarity of about $0.999$, and $\|\dv\|$ is only
$3.6\%$ of the full velocity (median; \cref{fig:anatomy:a}), too weak to compete with the FlowEdit velocity. We
therefore scale it by a constant $\lambda$ chosen on the analysis set. $\lambda\dv$ is the guidance term of
classifier-free guidance on the editor with $c^{\mathrm{neu}}$ as the negative prompt, here added to the velocity of a
different model. It raises signal and nuisance alike, so the amplifier is always combined with the three operators
below.

\paragraph{(b) Informative only at high noise $\rightarrow$ noise-level window.}
On latent frame~$0$, $\cos(\dv,-\mathbf{D})$ is $0.004$, $0.037$ and $0.083$ at $\sigma=0.3$, $0.5$ and $0.7$, and
rises to $0.058$, $0.125$ and $0.171$ after $8\times$ average pooling (\cref{fig:anatomy:b}). The edit direction is thus
carried by high noise levels and low spatial frequencies. At low noise, $\dv$ is dominated by fine-scale components
that would only damage details the video model has already resolved. We therefore apply the guidance only while
$\sigma \ge \sigma_{\mathrm{win}}$. The coarse appearance of the edit (color, material, identity) is settled in the
guided steps, and the unguided late steps restore fine detail and resolve remaining inconsistencies. With
$\sigma_{\mathrm{win}}=0.6$, chosen on the analysis set, the editor is queried in the first 24 of our 40 editing
steps. The same profile appears along \method{}'s own sampling trajectory, where $\dv$ is evaluated on the target state
of \cref{eq:g} rather than on noised source latents. On FiVE-Bench pairs, $\cos(\dv,-\mathbf{D}^{(0)})$ averages
$0.11$ inside the window ($0.23$ for $\sigma\ge0.85$), against $0.016$ for shuffled pairs, with 44 of the 46 pairs
above that level. Below $\sigma\approx0.5$ it falls to the shuffled level.
\paragraph{(c) Not localized $\rightarrow$ object gate.}
Let $\mathbf{M}^\star$ be the edited object, segmented with SAM2~\cite{ravi2024sam2} from the object named in the
instruction. On average, a fraction $0.88$ of the energy of $\dv$ falls outside $\mathbf{M}^\star$, as for a spatially
uniform signal ($0.89$, $\sqrt{1-|\mathbf{M}^\star|/|\Omega|}$), whereas the editor's own residual $\mathbf{D}$ gives
$0.62$ (\cref{fig:anatomy:c}). The pattern holds for every video, and the magnitude of $\dv$ does not reveal the object
either (AUC $0.58$). Amplified, $\dv$ perturbs the background as much as the object, which explains the blurred
backgrounds of naive guidance. Localization must therefore come from outside $\dv$: we gate it with a soft mask
$\mathbf{M}\in[0,1]^{K\times h\times w}$. Qwen2.5-VL~\cite{bai2025qwen25vl}, the text encoder already loaded with the
editor, extracts an object phrase $o$ from the instruction, and
CLIPSeg~\cite{luddecke2022clipseg} segments $o$ in every source frame. The probabilities are max-pooled over the RGB
frames of each latent frame and ($8\times$) over space, thresholded at $0.3$, dilated by two latent pixels and softened
with a $3\times3$ box filter. A coarse mask suffices, since within the dilated region the video model still decides
where exactly the edit lands; outside it, the update reduces to plain FlowEdit and inherits its background fidelity.

\paragraph{(d) Not temporally coherent $\rightarrow$ temporal smoothing.}
Because the editor sees each latent frame on its own, its edit velocities are nearly uncorrelated across frames: their
cosine similarity is $0.009\pm0.022$ between the first two latent frames and $0.04$ on average over all pairs at
$\sigma=0.7$, whereas the residuals $\mathbf{D}$ are clearly shared ($0.32$; \cref{fig:anatomy:d}). The intended edit
is the same in every frame, so the frame-specific remainder of $\dv$ behaves as temporally independent noise and,
amplified, produces flicker. We suppress it by averaging adjacent latent frames,
\begin{equation}
  \bar{\mathbf{g}}^{(k)}_\sigma = \tfrac14\,\mathbf{g}^{(k-1)}_\sigma + \tfrac12\,\mathbf{g}^{(k)}_\sigma + \tfrac14\,\mathbf{g}^{(k+1)}_\sigma,
  \label{eq:smooth}
\end{equation}
with replicate padding at both ends. A direction shared by neighboring frames passes through unchanged, whereas
uncorrelated components keep only $\tfrac{3}{8}$ of their energy. The kernel is intentionally short: a longer one
would smear the edit along the motion of the object.
\begin{figure*}[t]
\centering
\includegraphics[width=\linewidth]{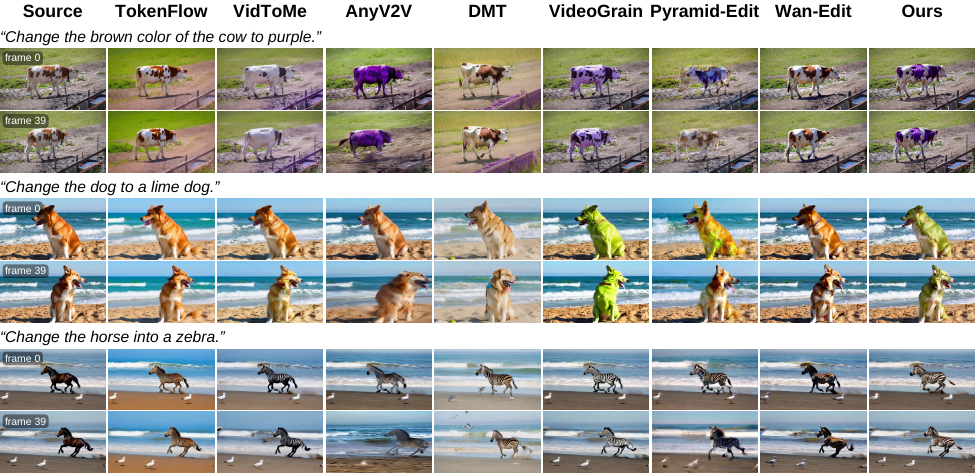}
\caption{\textbf{Qualitative comparison between \method{} and other baselines on FiVE-Bench.}
Each example shows frames 0 and 39 of a source video edited with the instruction above it.
Wan-Edit, which uses the same video model and sampler as \method{} but no edit velocity, makes weak or no change. Other baselines apply the edit only partially or alter the background over time. \method{} applies the edit in both frames while keeping the background and motion of the source.}
\label{fig:qual_five}
\end{figure*}

\paragraph{Full operator.}
\begin{equation}
  \begin{split}
  \mathcal{G}(\mathbf{g}_\sigma) = {}&\underbrace{\mathbf{1}[\sigma \ge \sigma_{\mathrm{win}}]}_{\text{window}}\;
  \underbrace{\lambda}_{\text{amplifier}}\; \underbrace{\mathbf{M}}_{\text{gate}}\odot
  \underbrace{\bar{\mathbf{g}}_\sigma}_{\text{smoothing}}
  \end{split}
  \label{eq:G}
\end{equation}
All four operators are linear in $\mathbf{g}_\sigma$, so their order does not matter. Even after $\mathcal{G}$, the
guidance is computed frame by frame and is far from a coherent video edit. It does not need to be one: it only biases
the coarse content of each latent frame at high noise, and the video model, which produces all frames jointly,
reconciles these biases into a single consistent appearance, while the unguided low-noise steps synthesize the
details. \method{} thus uses the image editor for \emph{what} to change and the video model for \emph{how} the change
evolves over time.

%% file: sec/experiment.tex
\section{Experiments}
\label{sec:exp}

\subsection{Setup}
\label{sec:exp:setup}

\paragraph{Models and hyper-parameters.}
\method{} combines Wan2.1-T2V-1.3B~\cite{wan2025} with Qwen-Image-Edit~\cite{wu2025qwenimage}; nothing is trained or
fine-tuned. One set of hyper-parameters is used for all videos and edit types: a 50-step schedule with timestep
shift~3, starting at step~10 ($\sigma\approx0.92$), CFG scales of 3 (source) and 7 (target), $\lambda=8$ and
$\sigma_{\mathrm{win}}=0.6$.

\paragraph{Data.}
We evaluate on FiVE-Bench~\citep{li2025five}: 100 videos with 419 edit pairs in six categories (rigid and non-rigid
replacement, color, material, addition and removal). Following its Wan-Edit protocol, we use the first 41 frames at
$832\times480$ and the provided prompts and instructions. FiVE's object masks are used only for evaluation. The
analysis of \cref{sec:method:anatomy} and the choice of $\lambda$ and $\sigma_{\mathrm{win}}$ use a separate set of 8
DAVIS-2017 videos~\citep{pont2017davis}.

\paragraph{Metrics.}
We report the official FiVE-Bench metrics: structure distance; background PSNR, LPIPS and SSIM outside the edit mask;
whole-frame and edited-region CLIP similarity; motion fidelity; and FiVE-Acc, the fraction of edits that a VLM judge
(Qwen2.5-VL-7B) deems successful. Since these are computed on every eighth frame and do not isolate blur, the
ablations also report the Laplacian sharpness inside and outside the edit mask, relative to the source, and the
RAFT~\citep{teed2020raft} warping error of consecutive frames under the source flow, which measures flicker.

\paragraph{Baselines.}
We compare with the training-free methods TokenFlow~\citep{geyer2024tokenflow}, VidToMe~\citep{li2024vidtome},
AnyV2V~\citep{ku2024anyv2v}, DMT~\citep{yatim2024dmt}, VideoGrain~\citep{yang2025videograin},
FlowDirector~\cite{li2026flowdirector}, Pyramid-Edit and Wan-Edit~\citep{li2025five}, and with
VACE~\cite{jiang2025vace}. Wan-Edit applies FlowEdit (\cref{eq:flowedit}) to the same
video model and is the most direct reference for what the edit velocity adds.

\subsection{Results}
\label{sec:exp:results}

\subsubsection{Qualitative Results}

\Cref{fig:qual_five} compares \method{} with Wan-Edit on FiVE-Bench. Both use the same video model and the same FlowEdit
mechanism. Wan-Edit, however, must infer the edit from the difference between two captions, which a text-to-video model
translates into weak or no change: the cat stays a cat and the person keeps their clothes. With the edit velocity of the
image editor, the same sampler turns the cat into a ferret , and changes color and material.
The edits are consistent from the first to the last frame, while the background, the camera and the motion follow the
source.
\subsubsection{Quantitative Results}

\begin{table*}[t]
\centering
\small
\setlength{\tabcolsep}{2.8pt}
\caption{\textbf{Comparison on FiVE-Bench}~\citep{li2025five}. Background preservation is measured outside the edit
mask; CLIP$_e$: text--video similarity on the edited region; MFS: motion fidelity; FiVE-Acc: VLM-judged edit success
(\%, official judge Qwen2.5-VL-7B); Acc$_2$: same protocol with Idefics3-8B; Time: s/frame on one
RTX PRO 6000. Struct, LPIPS $\times10^3$; SSIM, MFS $\times10^2$. Best in bold, second underlined.}
\label{tab:five}
\resizebox{\textwidth}{!}{%
\begin{tabular}{lcccccccccc}
\toprule
 & \multicolumn{1}{c}{Structure} & \multicolumn{3}{c}{Background preservation} & \multicolumn{2}{c}{Text alignment} & Motion & \multicolumn{2}{c}{Edit success} & Speed \\
\cmidrule(lr){2-2}\cmidrule(lr){3-5}\cmidrule(lr){6-7}\cmidrule(lr){8-8}\cmidrule(lr){9-10}\cmidrule(lr){11-11}
Method & Struct$\downarrow$ & PSNR$\uparrow$ & LPIPS$\downarrow$ & SSIM$\uparrow$ & CLIP$\uparrow$ & CLIP$_e\uparrow$ & MFS$\uparrow$ & FiVE-Acc$\uparrow$ & Acc$_2\uparrow$ & Time (s/frame)$\downarrow$ \\
\midrule
TokenFlow~\cite{geyer2024tokenflow} & 35.62 & 19.06 & 263.61 & 72.51 & 26.46 & 21.15 & 89.00 & 27.43 & 31.32 & 6.95 \\
VidToMe~\cite{li2024vidtome} & 22.37 & 21.15 & 263.91 & 70.69 & 26.84 & 21.05 & \textbf{90.06} & 26.77 & 29.29 & 1.94 \\
AnyV2V~\cite{ku2024anyv2v} & 71.36 & 15.90 & 348.59 & 50.77 & 24.89 & 19.72 & 60.36 & 38.02 & 35.14 & 12.36 \\
DMT~\cite{yatim2024dmt} & 85.95 & 14.71 & 404.60 & 51.64 & 26.66 & 21.44 & 82.30 & 48.42 & 44.30 & 13.82 \\
VideoGrain~\cite{yang2025videograin} & \textbf{12.40} & \textbf{27.05} & 185.21 & 79.13 & 25.69 & 20.31 & 88.57 & 37.23 & 49.77 & 16.47 \\
Pyramid-Edit~\cite{li2025five} & 28.65 & 20.84 & 276.59 & 71.72 & 26.82 & 20.20 & 80.59 & 43.84 & 45.44 & \textbf{0.39} \\
Wan-Edit~\cite{li2025five} & \underline{12.53} & \underline{25.57} & \underline{94.61} & \underline{82.55} & 26.39 & 21.23 & \underline{89.43} & 46.97 & 46.03 & \underline{1.79} \\
FlowDirector~\cite{li2026flowdirector} & 20.13 & 22.76 & 102.41 & 82.32 & 26.99 &21.15 & 87.89 & 56.38 & 54.13 & 8.40 \\
VACE~\cite{jiang2025vace} & 38.46 & 18.79 & 164.51 & 63.73 & \underline{27.03} & \textbf{21.78} & 83.26 & \underline{59.17} & \underline{60.11} & 10.39 \\
\midrule
\method{} (ours) & 13.42 & 25.03 & \textbf{89.17} & \textbf{83.09} & \textbf{27.06} &  \underline{21.47} & 88.83 & \textbf{65.60} & \textbf{60.45} & 8.01 \\
\bottomrule
\end{tabular}%
}
\end{table*}

\Cref{tab:five} reports the official FiVE-Bench metrics. \method{} achieves the highest
edit success (FiVE-Acc \textbf{65.6}), exceeding VACE, which is trained on video data, by
6.4 points, the best training-free baseline, FlowDirector, by 9.2, and Wan-Edit, which edits
the same backbone without the edit velocity, by 18.6. Against Wan-Edit, the gain holds on all
four FiVE-Acc criteria and pair by pair. Our
sampler alone reaches only 41.9 (\cref{tab:ablation}), so the gain comes from the editor, not
the sampler settings. Fidelity is preserved: \method{} attains the best LPIPS, SSIM and
whole-frame CLIP, with background PSNR close to Wan-Edit (25.0 vs.\ 25.6). Only VideoGrain
preserves the background better (27.1), at far lower edit success (37.2). Structure distance
and motion fidelity trail Wan-Edit slightly (13.4 vs.\ 12.5; 88.8 vs.\ 89.4).

Over Wan-Edit, the gains are largest for appearance
edits (color $64\!\to\!90$, material $33\!\to\!51$), followed by rigid ($54\!\to\!70$) and
non-rigid replacement ($50\!\to\!64$). Removal rises from 0 to 35, though addition and removal
contain only 9 and 10 pairs. On a 59-pair subset, editing each frame independently raises edit
success (84.9 vs.\ 63.9) but loses temporal coherence: warping error rises from 6.5 to 9.5,
motion fidelity drops from 88.9 to 82.5, and background PSNR from 25.1 to 21.0.

Because the official judge shares a model family with our editor, we repeat the evaluation
with Idefics3-8B~\citep{laurencon2024idefics3} under the same protocol (Acc$_2$). \method{}
again ranks first (60.5), on par with VACE (60.1) and well ahead of FlowDirector (54.1) and
Wan-Edit (46.0). On the four categories VideoGrain covers, it scores 67.6 vs.\ 49.8.

\begin{table}[t]
\centering
\footnotesize
\setlength{\tabcolsep}{3pt}
\caption{\textbf{Swapping the editor (a) or the video model (b, c) without retuning.}
Defaults fixed ($\lambda=8$, $\sigma_{\mathrm{win}}=0.6$, CFG 7/3). ``none'': same video model without an editor
(FlowEdit); $\Delta$: FiVE-Acc gain over it. QIE/QIE-2511: Qwen-Image-Edit-2509 (default)/2511;
Chrono: ChronoEdit-14B~\cite{wu2025chronoedittemporalreasoningimage}.}
\label{tab:app_swap}
\begin{tabular}{@{}l ccc ccc c@{}}
\toprule
& \multicolumn{3}{c}{Fidelity} & \multicolumn{3}{c}{Editing} & \\
\cmidrule(lr){2-4}\cmidrule(lr){5-7}
Editor & Str.$\downarrow$ & PSNR$\uparrow$ & LPIPS$\downarrow$ & CLIP$_e\uparrow$ & MFS$\uparrow$ & Acc$\uparrow$ & $\Delta$ \\
\midrule
\multicolumn{8}{@{}l}{\textit{(a) Wan2.1-T2V-1.3B}} \\
none     & 9.5  & 26.3 & 93.0 & 20.8 & 92.1 & 35.4 & -- \\
QIE      & 14.4 & 25.2 & 98.7 & 21.3 & 92.0 & 62.5 & \textbf{+27.1} \\
QIE-2511 & 14.5 & 25.2 & 98.6 & 21.4 & 91.2 & 63.5 & \textbf{+28.1} \\
Chrono   & 11.6 & 25.7 & 95.8 & 21.1 & 91.7 & 51.0 & \textbf{+15.6} \\
\midrule
\multicolumn{8}{@{}l}{\textit{(b) Wan2.1-T2V-14B}} \\
none     & 10.0 & 26.7 & 92.6 & 21.0 & 92.7 & 38.5 & -- \\
QIE      & 13.9 & 25.4 & 99.1 & 21.0 & 91.3 & 63.5 & \textbf{+25.0} \\
\midrule
\multicolumn{8}{@{}l}{\textit{(c) Wan2.2-T2V-A14B}} \\
none     & 15.9 & 24.4 & 114.2 & 21.4 & 91.8 & 46.9 & -- \\
QIE      & 18.6 & 23.6 & 121.4 & 21.4 & 91.0 & 63.5 & \textbf{+16.7} \\
\bottomrule
\end{tabular}
\end{table}

\paragraph{Generality across models sharing a VAE.}
\method{} only assumes that the editor and the video model predict velocities in the same latent space
(\cref{sec:method:latent}), so any model built on the Wan2.1 VAE should be a drop-in replacement. We test this, not
a higher score, by swapping in Qwen-Image-Edit-2511, ChronoEdit-14B (fine-tuned from Wan2.1-I2V, without a
vision-language model), Wan2.1-T2V-14B and the mixture-of-experts Wan2.2-T2V-A14B, keeping all hyper-parameters at
their defaults. Every combination improves FiVE-Acc over
the same video model without an editor (\cref{tab:app_swap}), by $+15.6$ to $+28.1$ points. Stronger video models already edit better on their own
($35.4$, $38.5$ and $46.9$ for Wan2.1-1.3B, Wan2.1-14B and Wan2.2-A14B), so the gain shrinks, but with QIE the final
FiVE-Acc stays at $62.5$--$63.5$ on all three. The varying gains reflect untuned settings: ChronoEdit's edit velocity
is several times weaker than QIE's, so $\lambda=8$ under-amplifies it, and Wan2.2-A14B already preserves less than
Wan2.1 without an editor (PSNR $24.4$ vs.\ $26.3$), suggesting that the CFG scales tuned for Wan2.1 do not suit it.
These results are therefore conservative estimates.

\subsection{Why the Cross-Model Composition Works}
\label{sec:exp:analysis}

We analyze 48 randomly drawn FiVE-Bench pairs, used only for this analysis; the hyper-parameters remain those chosen
on the DAVIS analysis set. All quantities are measured along \method{}'s own sampling trajectory.

\paragraph{The editor supplies a direction.}
On the 46 pairs with a non-empty gate, $\cos(\dv,-\mathbf{D}^{(0)})$ is $0.11$ inside the noise window, far above a
shuffled-pair null ($0.016$), and vanishes below it ($-0.003$).
At the same time, $\dv$ is nearly uncorrelated with the video velocity ($-0.06$) and with the FlowEdit update
($0.02$), so it does not duplicate what the video model already does.
With $\lambda=8$ and the gate, the norm of $\mathcal{G}(\dv)$ reaches about two thirds of that of $\Delta_\sigma$
inside the window, so the two terms are of the same order.

\paragraph{The gain is semantic.}
An alternative explanation is that \emph{any} perturbation of the video velocity changes the output.
\Cref{tab:direction} rules this out. Each control is matched in norm to $\dv$ per frame and step and passes through
the same $\mathcal{G}$. Neither noise nor the edit velocity of another pair or of a wrong instruction changes
FiVE-Acc by more than 2.1 points, and all confidence intervals include zero, although the last two move the trajectory
nearly as far as the correct $\dv$. Reversing the edit velocity lowers FiVE-Acc to 17.7. The video model thus follows the direction that the editor provides, not merely its magnitude.

\begin{table}[t]
\centering
\footnotesize
\setlength{\tabcolsep}{3pt}
\caption{\textbf{Is the gain semantic?} Every injected signal is matched in norm to $\dv$ per
frame and step and passes through the same $\mathcal{G}$. $\Delta$: FiVE-Acc change vs.\ no injection with 95\%
 bootstrap CI; W/L: pairs more / less successful; $D_{\mathrm{traj}}$: deviation from the video-only trajectory.}
\label{tab:direction}
\begin{tabular}{lcccc}
\toprule
Injected signal & Acc$\uparrow$ & $\Delta$ [95\% CI] & W/L & $D_{\mathrm{traj}}$ \\
\midrule
None (video model only) & 36.5 & -- & -- & 0 \\
$\dv$ (\method{}) & \textbf{62.5} & $+26.0$ [16.7, 36.5] & 18/0 & 0.21 \\
\midrule
White noise & 34.4 & $-2.1$ [$-6.3$, 2.1] & 1/3 & 0.03 \\
Coarse-scale noise & 37.5 & $+1.0$ [$-3.1$, 5.2] & 3/2 & 0.05 \\
$\dv$ of another pair & 37.5 & $+1.0$ [$-4.2$, 6.3] & 4/3 & 0.15 \\
$\dv$ of a wrong instruction & 37.5 & $+1.0$ [$-5.2$, 7.3] & 5/3 & 0.17 \\
$-\dv$ & 17.7 & $-18.8$ [$-29.2$, $-9.4$] & 1/13 & 0.15 \\
\bottomrule
\end{tabular}
\end{table}

\begin{figure}[t]
\centering
\includegraphics[width=\linewidth]{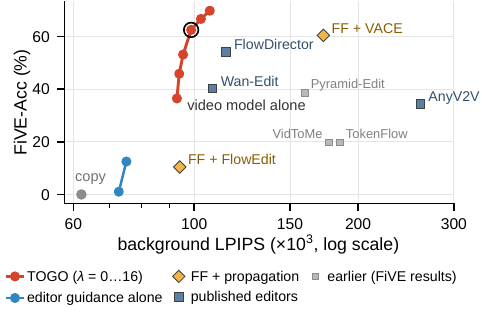}
\caption{\textbf{Edit success vs.\ background preservation} \method{} for $\lambda\in\{0,\dots,16\}$ ($\lambda=0$: video model alone; default $\lambda=8$ circled), the
editor's guidance alone, copying the source, and the baselines on the same pairs. FF: first frame edited with the same
editor, then propagated.}
\label{fig:pareto}
\end{figure}

\paragraph{Both models are needed.}
Neither model edits on its own (\cref{fig:pareto}): the video model alone reaches a FiVE-Acc of 36.5, and the editor's
guidance without $\Delta_\sigma$ at most 12.5. Composed, FiVE-Acc rises monotonically with $\lambda$, from 45.8 at
$\lambda=2$ to 69.8 at $\lambda=16$, while background PSNR falls by at most 2.2\,dB and motion fidelity by 2.0 points;
at $\lambda=8$, \method{} is more successful than the video model alone on 18 pairs and less successful on none.
Among methods evaluated on all pairs, only first-frame propagation with the same editor edits about as often, and at a
much higher background LPIPS: FF + VACE is both less successful than the default \method{} (60.4 vs.\ 62.5) and less faithful (LPIPS 172.7
vs.\ 98.7).

\subsection{Ablation Study}
\label{sec:exp:ablation}

\begin{table}[h]
\centering
\footnotesize
\setlength{\tabcolsep}{3pt}
\caption{\textbf{Ablation of $\mathcal{G}$.} Top: components added one at a time; bottom: one removed from the full
model. Warp: RAFT warping error (source: 3.49). Sharp.: Laplacian variance relative to the source inside / outside
the edit mask.}
\label{tab:ablation}
\begin{tabular}{lccccc}
\toprule
Variant & Acc$\uparrow$ & PSNR$\uparrow$ & MFS$\uparrow$ & Warp$\downarrow$ & Sharp.\ in/out$\uparrow$ \\
\midrule
Video model only & 41.91 & 26.41 & 89.48 & 5.99 & 1.10 / 0.91 \\
\midrule
+ Amplifier & \textbf{74.18} & 20.79 & 78.85 & 8.53 & 0.87 / 0.61 \\
+ Window & \underline{70.73} & 21.40 & 82.37 & 6.99 & 1.07 / 0.67 \\
+ Gate & 66.78 & 24.82 & 88.55 & 6.37 & \textbf{1.22} / \textbf{0.90} \\
+ Smoothing (\textbf{Full}) & 65.60 & \textbf{25.03} & \textbf{88.83} & \textbf{6.18} & \underline{1.16} / \underline{0.89} \\
\midrule
Full $-$ Gate & 69.64 & 21.72 & 84.09 & 6.49 & 0.96 / 0.65 \\
Full $-$ Window & 68.38 & \underline{24.96} & \underline{88.72} & \underline{6.30} & 1.09 / \underline{0.89} \\
\bottomrule
\end{tabular}
\end{table}

\paragraph{Guidance operator.}
As predicted in \cref{sec:method:anatomy}, the amplifier alone gives the highest FiVE-Acc (74.2) but amplifies the
nuisance in $\dv$ together with the edit (\cref{tab:ablation}): background PSNR drops from 26.4 to 20.8, motion
fidelity from 89.5 to 78.9, and the background loses 39\% of its sharpness, which FiVE-Acc does not penalize. The
window removes most of the flicker and restores the sharpness of the edited object, the gate repairs the background
(removing it from the full model costs 3.3\,dB and doubles the structure distance), and smoothing further lowers the
warping error. The full model thus trades 8.6 points of FiVE-Acc for a fidelity close to that of the video model
alone, and the window also saves 40\% of the editor calls.

%% file: sec/conclusions.tex
\section{Conclusion}
\label{sec:conclusion}

We have shown that an instruction-based image editor can be turned into a video editor without training. Because the
two share a latent space, the editor's edit velocity can be added to the sampler of a video model, and a simple
operator that amplifies it and restricts it in noise level, space, and time is enough for the video model to turn the
per-frame guidance into a temporally coherent edit. On FiVE-Bench, \method{} reaches an edit success rate of 65.6\%,
6.4 points above the strongest baseline, while keeping background preservation and motion fidelity close to those of
the video model alone. More broadly, progress in image editing need not be repeated for video: the video editor is
assembled at inference time, and a newer image editor or a ten times larger video model can be swapped in without
retuning.